\documentclass[runningheads]{llncs}

\usepackage{eccv}

\usepackage{eccvabbrv}
\usepackage{svg}
\usepackage{graphicx}
\usepackage{booktabs}
\usepackage{array}
\usepackage{multirow}
\usepackage{multicol}
\usepackage[export]{adjustbox}
\usepackage[accsupp]{axessibility}  

\usepackage{hyperref}

\usepackage{orcidlink}

\begin{document}

\title{SplatPose+: Real-time Image-Based Pose-Agnostic 3D Anomaly Detection} 


\author{Yizhe Liu \orcidlink{0009-0004-9572-5203} \and
Yan Song Hu \orcidlink{0009-0004-6428-672X} \and 
Yuhao Chen \orcidlink{0000-0001-6094-0545} \and 
John Zelek \orcidlink{0000-0002-8138-3546}}

\authorrunning{Yizhe Liu, \etal}

\institute{University of Waterloo \\
\email{\{yizhe.liu,y324hu,yuhao.chen1,jzelek\}@uwaterloo.ca}}

\maketitle
\begin{abstract}
  Image-based Pose-Agnostic 3D Anomaly Detection is an important task that has emerged in industrial quality control. This task seeks to find anomalies from query images of a tested object given a set of reference images of an anomaly-free object. The challenge is that the query views (a.k.a poses) are unknown and can be different from the reference views. Currently, new methods such as  OmniposeAD and SplatPose have emerged to bridge the gap by synthesizing pseudo reference images at the query views for pixel-to-pixel comparison. However, none of these methods can infer in real-time, which is critical in industrial quality control for massive production. For this reason, we propose SplatPose+, which employs a hybrid representation consisting of a Structure from Motion (SfM) model for localization and a 3D Gaussian Splatting (3DGS) model for Novel View Synthesis. Although our proposed pipeline requires the computation of an additional SfM model, it offers real-time inference speeds and faster training compared to SplatPose. Quality-wise, we achieved a new SOTA on the Pose-agnostic Anomaly Detection benchmark with the Multi-Pose Anomaly Detection (MAD-SIM) dataset.  
  
\end{abstract}

\section{Introduction}
\label{sec:intro}
Inspection is a critical procedure in manufacturing to sort out defective products with manufacturing flaws such as missing parts, stains, burrs, and holes. Traditionally, humans are employed to visually inspect the products on assembly lines; however, because of the high cost and human errors, algorithms have been developed to automate anomaly detection by comparing photos of a standard product (reference images) and photos of a product to be tested (query images). Traditional approaches \cite{pang2021explainabledeepfewshotanomaly, Wan2022, Ding2022} rely on supervised training with both normal and defective samples. According to a survey on anomaly detection conducted by Liu, \etal \cite{Liu2024}, more recent anomaly detection methods have shifted towards unsupervised training, using only normal samples. The reason behind the shift towards unsupervised training is the dis-proportionality of positive and negative samples and the inability of supervised models to generalize on unseen anomalous samples. One challenge in developing an unsupervised algorithm is that the model only sees an anomaly-free object from a set of predetermined reference views, while, in the testing stage, the object may have arbitrary orientations. Thus, anomaly detection algorithms need to be able to detect defects from images taken from any viewing angle, which is unknown and can be different from the reference views. This specific problem has been researched and named by Zhou, \etal, as the Pose-Agnostic 3D Anomaly Detection (PAD) problem\cite{zhou2023paddatasetbenchmarkposeagnostic}.

Intuitively, this problem can be broken down into two sub-problems. First, the poses of query images are unknown. Second, no reference images are available for the query views. The first sub-problem can be solved by building a structure-from-motion (SfM) model to localize query images. However, before the recent development of Novel View Synthesis (NVS) methods, the second sub-problem is hard to solve. SfM models, though able to optimize a mesh for the scene, cannot render high-fidelity unseen views. 
Recent developments in learning-based Novel View Synthesis (NVS) methods including Neural Radiance Field (NeRF) and 3D Gaussian Splatting (3DGS) enable direct pixel-to-pixel comparison by synthesizing pseudo reference images from the query view. Current state-of-the-art (SOTA) PAD methods rely on the same learning-based 3D representation for both sub-problems: localization and NVS. However, localization via learning-based 3D representation is costly because these methods need to back-propagate gradients from the image reconstruction loss to the camera poses, and refine those poses through gradient descent hundreds of times. As a result, current methods could take seconds, or even minutes, to detect anomalies in a query image.

\begin{figure}
    \centering
    \includegraphics[width=0.8\linewidth]{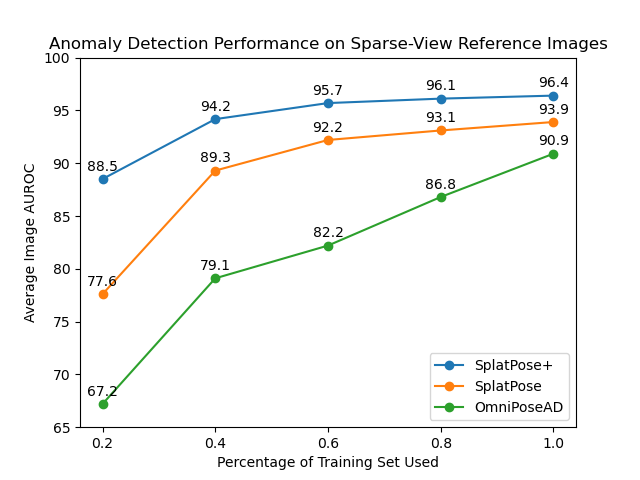}
    \includegraphics[width=0.8\linewidth]{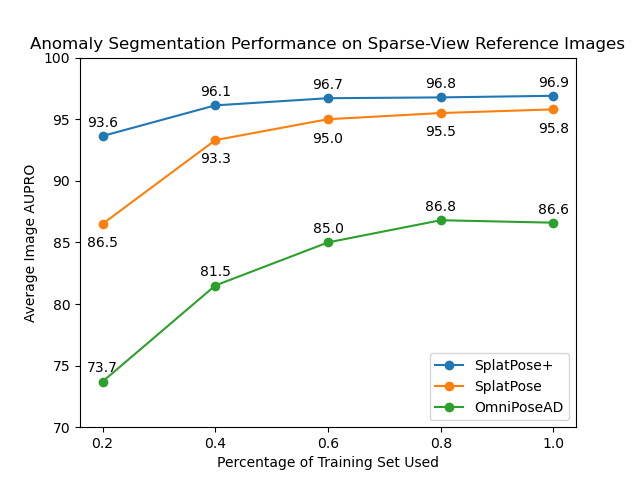}
    \caption{Quantitative Comparison for Anomaly Detection Performance on Sparse-View Reference Images}
    \label{fig:sparse_view}
\end{figure}

Therefore, in this work, we propose an unsupervised method: SplatPose+, to solve the PAD task in real time. SplatPose+ uses a hybrid 3D representation, combining a feature-matching-based SfM model and a learning-based 3DGS model. This hybrid structure enables us to efficiently localize query views and synthesize high-fidelity anomaly-free pseudo reference images for direct pixel-to-pixel comparison. Efficiency-wise, SplatPose+ is $37.4\%$ faster in training and $147$ times faster in inference than the current SOTA: SplatPose evaluated on the MAD-Sim dataset. Accuracy-wise, SplatPose+ reaches the new SOTA on the MAD-Sim dataset as shown in \cref{tab:img_auroc} and \cref{tab:pixel_scores}. Moreover, SplatPose+ outperforms current methods given sparse-view training data. As illustrated in \cref{fig:sparse_view}, when only using $40\%$ of the training data, the anomaly detection and segmentation scores of SplatPose+ are still higher than the current SOTA trained on all reference images. To summarize, our contributions include:
\begin{itemize}
    \item The first real-time image-based 3D pose-agnostic anomaly detection method. 
    \item Achieves top efficiency and best anomaly detection and segmentation scores on the MAD-Sim Dataset. 
    \item Robust anomaly detection performance given sparse-view training data.
    \item Code will be released: \url{https://github.com/Yizhe-Liu/SplatPosePlus}.
\end{itemize}

\section{Related Work}
\subsection{3D Reconstruction}
In computer vision, 3D Reconstruction is a process that computes a 3D representation from a set of 2D images of a scene or an object from multiple viewing angles. Traditionally, structure-from-motion (SfM) methods are developed to optimize a sparse model and extract poses from a set of images through feature extraction and feature matching. Colmap \cite{schoenberger2016SfM, schoenberger2016mvs} is a well-known SfM pipeline that uses SIFT \cite{martinezgil2016siftalgorithmextractingstructural} as the feature extractor and RANSAC \cite{Fischler1981} as the feature matcher. Later on, a more advanced and flexible SfM pipeline: Hierarchical Localization (hloc) \cite{sarlin2019coarsefinerobusthierarchical} has been proposed to incorporate more advanced feature extractors (SuperPoint \cite{detone2018superpointselfsupervisedpointdetection} and DISK \cite{tyszkiewicz2020disklearninglocalfeatures}) and feature matchers (SuperGlue \cite{wang2020supergluestickierbenchmarkgeneralpurpose} and LightGlue \cite{lindenberger2023lightgluelocalfeaturematching}), which can generate better feature descriptors and more robust matches. 

\subsection{Novel View Synthesis (NVS)}
Novel view synthesis is a task that synthesizes arbitrary views from a given set of images of a scene or an object. Although SfM methods can reconstruct dense 3D meshes, 3D meshes generally cannot generate high-fidelity images from unseen views. As a result, new classes of learning-based methods have been proposed for this task, including Neural Radience Field (NeRF) \cite{mildenhall2020nerfrepresentingscenesneural} and 3D Gaussian Splatting \cite{kerbl20233dgaussiansplattingrealtime}

\subsubsection{Neural Radiance Field (NeRF) \cite{mildenhall2020nerfrepresentingscenesneural}} represents a scene of an object using an implicit neural representation. NeRF deconstructs each image into individual pixels, with each pixel representing a unique ray that extends from the camera to the object in the scene. A set of 5D Points, consisting of coordinates ($x, y, z$) for the location and ($\theta, \phi$) for the viewing angle, are sampled along each ray as input for a multilayer perceptron (MLP) that predicts the color and density of each point. Finally, the color and density values predicted along a ray are accumulated via volume metric rendering to calculate the pixel color. NeRF is able to synthesize photorealistic novel views. However, NeRF is computationally expensive, taking hours to train for each scene. This is because it is an implicit representation, which requires millions of queries of the MLP to generate a single image. 

\subsubsection{3D Gaussian Splatting \cite{kerbl20233dgaussiansplattingrealtime}}, unlike NeRF, represents a scene or an object using an explicit representation consisting of anisotropic 3D Gaussian colorized by spherical harmonics. 3D Gaussians are initialized by randomization or a reference sparse point cloud produced by a SfM system. Rays are sampled in a similar fashion as in NeRF. However, the color of a ray is accumulated by the view-dependent color and density values from 3D Gaussians that the ray intersects. By using an explicit representation, no MLP queries are required. As a result, 3DGS can be trained within minutes and rendered in real time. 

\subsection{Image-Based Pose-Agnostic 3D Anomaly Detection}
\subsubsection{OmniposeAD \cite{zhou2023paddatasetbenchmarkposeagnostic}} is the first Pose-Agnostic 3D Anomaly Detection method proposed together with the MAD-Sim dataset. It utilizes NeRF to reconstruct pseudo reference and query image pairs for pixel-to-pixel comparison. A neural radiance field is trained using multiview images from the anomaly-free reference object. For each query image, the coarse query pose is determined by the pose of the reference image that has the most LoFTR \cite{sun2021loftrdetectorfreelocalfeature} matches with the query image. INeRF \cite{yenchen2021inerfinvertingneuralradiance} optimizes the coarse pose for the fine pose given the reference NeRF model. Then, NeRF synthesizes the pseudo reference image as if captured from the query view using the fine pose localized from each query image. After that, a pre-trained CNN is applied to extract features from the pseudo reference and query image pair. Finally, the anomaly scores are calculated by the L2 difference between the feature maps and the difference in RGB values. According to Zhou, \etal, OmniposeAD outperformed other non-NVS-based methods \cite{zhou2023paddatasetbenchmarkposeagnostic} at that time. However, OmniposeAD is slow to train ($4.5$ hours) and infer ($1$ minute). 

\subsubsection{SplatPose\cite{kruse2024SplatPosedetectposeagnostic}} is proposed to improve the efficiency and the anomaly detection accuracy of OmniposeAD \cite{zhou2023paddatasetbenchmarkposeagnostic}. A well-known issue for NeRF is the prohibitively high training and inference time. According to Kruse, \etal. \cite{kruse2024SplatPosedetectposeagnostic}, SplatPose reduce the training and inference time by $2$ magnitudes, which is achieved by replacing NeRF and INeRF with a pose-wise differentiable 3D Gaussian Splatting implementation. Moreover, the qualitative anomaly detection scores, and sparse-view robustness (defined in \cref{section_sparse_view}) are also improved thanks to the better NVS ability of 3DGS.

\section{Method}
For the pose-agnostic anomaly detection task, we define the reference set as $R = \{(r_1, T_1), (r_2, T_2), \cdots, (r_n, T_n)\}$ and the query set as $Q = \{q_1, q_2, \cdots, q_m\}$ where the pose information $T_i$ is known only for the reference set. The goal is to find image-level and pixel-level anomalies in $Q$ while only trained on $R$.

During the training stage, we first build a SfM model using hloc \cite{sarlin2019coarsefinerobusthierarchical} with the configuration of (NetVLAD \cite{arandjelović2016netvladcnnarchitectureweakly} + Superpoint \cite{detone2018superpointselfsupervisedpointdetection} + LightGlue \cite{lindenberger2023lightgluelocalfeaturematching}). Then, we use the sparse point cloud from the SfM model as the initialization to train a 3DGS model. 
During the testing stage, we first localize the query images against the SfM model to obtain the query poses. Then, the 3DGS model synthesizes pseudo reference images at those query poses. Lastly, a pre-trained CNN is employed to calculate the anomaly score. 

\begin{figure}[ht]
  \centering
  \includesvg[width=0.9\linewidth]{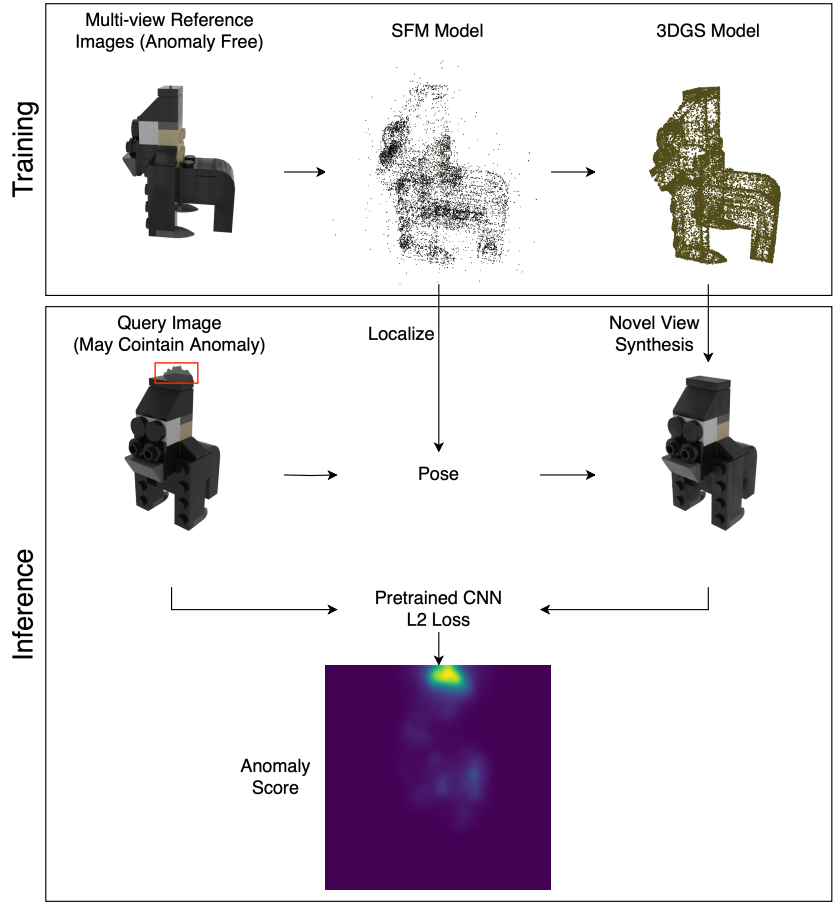}
  \caption{SplatPose+ pipeline incorporates an SfM model for localizing query views and initialization for the 3DGS model. 3DGS model is trained only for Novel View Synthesis.}
  \label{fig:pileline}
\end{figure}

The current SOTA: SplatPose uses the same 3DGS for both localization and NVS. In comparison, our method triangulates an additional SfM model for 3DGS initialization and localization, and the hybrid representation (3DGS and SfM) can train faster and detect anomalies better in real-time. The pipeline is illustrated in \cref{fig:pileline}.  

\subsection{Hybrid Representation}
\subsubsection{SfM Model}
We employ hloc \cite{sarlin2019coarsefinerobusthierarchical}, which is the SOTA modularized visual localization pipeline, to build the SfM model. For the training stage, both images and poses are known for the reference views. We use NetVLAD \cite{arandjelović2016netvladcnnarchitectureweakly}, a lightweight but effective image-level feature descriptor for image retrieval, to build a reference database $D^R$. Then, overlapped reference image pairs are extracted from $D^R$. After that, we use SuperPoint \cite{detone2018superpointselfsupervisedpointdetection} to extract local features and use LightGlue \cite{lindenberger2023lightgluelocalfeaturematching} to match features. We choose SuperPoint and LightGlue because SuperPoint can find more feature points than traditional methods, for example, SIFT, and LightGlue is the SOTA lightweight feature matcher. The matched features are then triangulated to build a reference sparse model $M^s$.
For the inference stage, the poses for the query views are unknown. To localize query views, we also use NetVLAD for the image retrieval against the reference database $D^R$ to find $15$ most related reference images for each query image. NetVLAD helps speed up both training and inference by narrowing down the search space because we only need to match features between a fixed number of similar images instead of an exhaustive search. Similarly, SuperPoint features are extracted from matched reference and query image pairs. Then, LightGlue is employed to match features. Finally, the query views are localized against the sparse model $M^s$ from the matched features by hloc. 

\subsubsection{Novel View Synthesis} We first train a 3DGS model and then use the poses localized by the SfM model $M^s$ to reconstruct pseudo reference images at the query views. Unlike SplatPose, which trains a 3DGS model from random initialization, SplatPose+ trains a 3DGS model $M^{gs}$ initialized by the sparse point cloud extracted from the sparse SfM model $M^s$, in which points are resided and concentrated on the true object surfaces. Moreover, SplatPose only uses RGB to colourize gausssians (maximum spherical harmonics degree: SHS $=0$) in 3DGS for stable gradient descent during pose optimization. Since we no longer use 3DGS for localization, we are able to use the full spherical harmonics degree of $3$ for better view-dependent colourization. Lastly, inspired by the experiments conducted in Compact3D \cite{navaneet2024compact3dsmallerfastergaussian}, we reduce the densification frequency by setting the densification interval to $1000$ because the SfM point cloud is already well-distributed on the object surface illustrated in \cref{fig:pileline}. With good initialization, increasing the densification interval greatly improves the efficiency while causing marginal deterioration in reconstruction quality. Our experiments \cref{tab:3dgs_comp} show that with those three changes, our 3DGS model trains $2$x faster than SplatPose and achieves comparable construction quality quantified by the peak signal-to-noise ratio (PSNR).

\begin{table}[ht]
    \centering
    \begin{tabular}{l|c|c|c|c|c|c}
    \toprule
    3DGS Setting & Initialization & Iters & Densification & SHS & Avg. PSNR & Train Time (s)\\
    \midrule 
    SplatPose \cite{kruse2024SplatPosedetectposeagnostic} & Random & 30k & 100 & 0 & 42.13 & 294 \\
    SplatPose+ & SfM & 15k & 1000 & 3 & 42.12 & 107  \\
    \bottomrule
    \end{tabular}
    \caption{Reconstruction efficiency and quality comparison. SHS: Spherical Harmonics Degree. }
    \label{tab:3dgs_comp}
\end{table}

\subsection{Anomaly Score}
Given a query image $q$ and the pseudo reference image $\hat{r}$ generated by 3DGS from the same view (paired image), we used the same pre-trained $5$-layer Convolutional Neural Network (CNN) used in OmniPoseAD \cite{zhou2023paddatasetbenchmarkposeagnostic}, to extract $5$ features $(f_1^{\hat{r}},f_1^{q}), \cdots, (f_5^{\hat{r}}, f_5^q)$ from the both images from layer $1$ to layer $5$ with decreasing spatial resolution. We use the same CNN as well as the L2 norm to calculate the anomaly map for a fair comparison. Pixel-wise L2 norms are calculated between the paired reference and query image $\displaystyle ||q - \hat{r}||_2$ and all feature pairs $\displaystyle ||f_1^{\hat{r}} - f_1^{\hat{r}}||_2, \cdots, ||f_5^{\hat{r}} - f_5^{\hat{r}}||_2$. All L2 norms are then resized to $224*224$ and then summed, Gaussian-smoothed and normalized to form the final pixel-wise anomaly score. Image anomaly scores are assigned by the maximum pixel anomaly score in each image.

\section{Evaluation}
\subsection{Dataset}
\begin{figure}[ht]
  \centering
  \includegraphics[height=6.5cm]{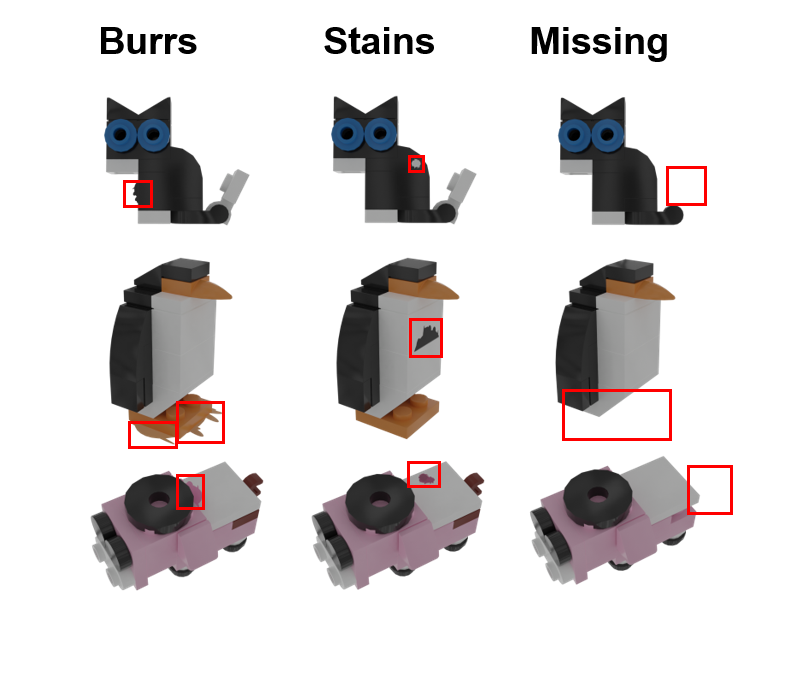}
  \caption{Defect Sample in MAD Dataset\cite{zhou2023paddatasetbenchmarkposeagnostic}}
  \label{fig:MAD}
\end{figure}

The Simulated Multi-pose Anomaly Detection (MAD-Sim) dataset is the only dataset for the PAD task proposed together by Zhou, \etal
\cite{zhou2023paddatasetbenchmarkposeagnostic}. The MAD-Sim dataset, illustrated in \cref{fig:MAD}, contains $3$ types of defects (Burrs, Stains and Missing Parts) for $20$ different synthetic Lego objects. This dataset mimics the standardized product defect detection scenario in the real world where products to be tested can have arbitrary orientation when passing a checkpoint with multiple fixed cameras. MAD-Sim is an object-central anomaly detection dataset. For each object, MAD-Sim provides \begin{itemize}
    \item $210$ pose-known reference images (without defects) from different views.
    \item Approximately $300$ pose-unknown query images from different views with different types of defects, and the masks of the defects. 
\end{itemize}

\subsection{Metrics}
We use the well-established metrics AUROC and AUPRO \cite{Bergmann2021} to evaluate the anomaly detection and segmentation of our model. AUROC evaluates the performance of a classification model for all thresholds by measuring the area under the receiver operating characteristic (ROC) curve. ROC curve plots true positive rates ($R_{\text{TP}}$) against false positive rates ($R_{\text{FP}}$) at all thresholds. A higher AUROC means the ROC curve bends further toward the top left, that is, the model classifies anomalies better by maintaining a higher $R_{\text{TP}}$ and a lower $R_{\text{FP}}$. 
\begin{equation}\text{AUROC} = \int R_{\text{TP}} \, d{R_{\text{FP}}}\end{equation}

AUPRO evaluates the performance of a segmentation model by measuring the area under the per-region-overlap (PRO) curve. PRO better describes the model's ability to segment anomalies of all sizes than intersection-over-union (IOU) because this metric calculates the overlap ratio per region, which weighs smaller anomalies the same as larger anomalies.

\begin{equation}
\text{PRO} = \frac{1}{N} \sum_{i=1}^N \frac{|P \cap C_i|}{|C_i|}
\end{equation}
\begin{equation}
\text{AUPRO} = \int \text{PRO} \, d{R_{\text{FP}}}
\end{equation}
where $P$ represents pixels classified as anomalies, and $C_i$ represents the set of pixels in the $i$-th connected anomalous region in the ground truth.

\subsection{Results}
We conduct all experiments under the unsupervised setting, where only anomaly-free reference images are used for training. For the SfM model, we use the default settings for NetVLAD, SuperPoint and LightGlue with num\_matched $=15$, that is, each query image is matched with $15$ most similar reference images using the NetVLAD features. For the 3DGS model, compared to SplatPose, we decrease the densification frequency from once every $100$ iterations to once every $1000$ iterations and the total training iterations from $30000$ to $15000$. 

Under these settings, our methods perform better than other NVS-based methods on image-level anomaly detection \cref{tab:img_auroc} as well as pixel-wise anomaly segmentation tasks \cref{tab:pixel_scores} on the MAD-SIM dataset. SplatPose+ achieved a higher image-level AUROC for $18$ out of $20$ test objects. Overall, SplatPose+ achieved image-level AUROC of $96.4$ and pixel-level AUPRO of $96.9$ on the PAD task, the new SOTA on the MAD-SIM dataset. 

\begin{table}[ht]
    \centering
    \begin{tabular}{l|c|c|c}
    \toprule
    Methods & OmniposeAD & SplatPose & SplatPose+ \\
    Objects & \cite{zhou2023paddatasetbenchmarkposeagnostic} & \cite{kruse2024SplatPosedetectposeagnostic} &  \\
    \midrule
    Gorilla  &  \underline{93.6} & 91.7 & \textbf{95.5} \\
    Unicorn  & 94.0 &  \underline{97.9} &\textbf{99.0}\\
    Mallard  & 84.7 &  \underline{97.4} & \textbf{98.3}\\
    Turtle  & 95.6 &  \underline{97.2} & \textbf{97.4} \\
    Whale  & 82.5 &  \underline{95.4} & \textbf{98.9} \\
    Bird  & 92.4 &  \underline{94.0} & \textbf{98.7} \\
    Owl  & \textbf{88.2} & 86.8 &  \underline{87.9} \\
    Sabertooth &  \underline{95.7} & 95.2 & \textbf{98.3} \\
    Swan  & 86.5 &  \underline{93.0} & \textbf{95.7}\\
    Sheep  & 90.1 &  \underline{96.7} & \textbf{97.4}\\
    Pig & 88.3 &  \underline{96.1} & \textbf{98.4}\\
    Zalika  & 88.2 &  \underline{89.9} & \textbf{93.9}\\
    Phoenix  & 82.3 &  \underline{84.2} & \textbf{90.4}\\
    Elephant  & 92.5 &  \underline{94.7} & \textbf{96.8}\\
    Parrot  & \underline{97.0} & 96.1 & \textbf{98.9}\\
    Cat  & \underline{84.9} & 82.4 & \textbf{89.4} \\
    Scorpion  & 91.5 & \underline{99.2} & \textbf{99.6}\\
    Obesobeso  & \textbf{97.1} & 95.7 & \underline{96.8}\\
    Bear  & 98.8 & \underline{98.9} & \textbf{99.3}\\
    Puppy  & 93.5 & \underline{96.1} & \textbf{98.0}\\
    \midrule
    Avg. & 90.9 & \underline{93.9} & \textbf{96.4}\\
    \bottomrule
    \end{tabular}
    \caption{Comparison of NVS-based methods on MAD-SIM using Image-level AUROC (The higher the better $\uparrow$). The best result per class is \textbf{bold} and the runner-up is \underline{underlined}.}
    \label{tab:img_auroc}
\end{table}

\begin{table}[ht]
    \centering
    \begin{tabular}{l|c|c|c}
    \toprule
    Methods & OmniposeAD & SplatPose & SplatPose+ \\
    Metrics & \cite{zhou2023paddatasetbenchmarkposeagnostic} & \cite{kruse2024SplatPosedetectposeagnostic} &  \\
    \midrule
    AUROC ($\uparrow$) & 98.4 & \underline{99.5} & \textbf{99.6} \\
    AUPRO ($\uparrow$) & 86.6 & \underline{95.8} & \textbf{96.9} \\ 
    \bottomrule
    \end{tabular}
    \caption{Pixel-wise anomaly segmentation on MAD. The best result per class is \textbf{bold} and the runner-up is \underline{underlined}.}
    \label{tab:pixel_scores}
\end{table}

\subsection{Efficiency}
Inference efficiency is crucial in industrial anomaly detection applications because each product usually has less than a second to be inspected on an assembly line. The training time is relatively unimportant because it is considered as a one-time overhead. Previous methods OmniposeAD and SplatPose do not have real-time inference capability. Our method can not only infer in real-time but also train faster. For a fair comparison, all experiments are conducted on Ubuntu using a single NVIDIA RTX 3090. For the training stage, SplatPose+ takes an average of $76.7$ seconds to build an SfM model and $107.2$ seconds to train a 3DGS model. For the inference stage, SplatPose+ takes an average of $22$ milliseconds on localization, $9.8$ milliseconds on novel view synthesis, and $2.5$ milliseconds on CNN feature extraction and anomaly score calculation. Overall, on average, the training time per object is only $184$ seconds and the inference time is $34$ milliseconds per query view. Equivalently, the inference can be run in real-time at approximately $30$fps.

\begin{table}[ht]
  \label{tab:efficiency}
  \centering
  \begin{tabular}{ l | c | c }
    \toprule
    Method & Training (h:min:sec) $\downarrow$ & Inference (sec) $\downarrow$ \\
    \midrule
    OmniposeAD \cite{zhou2023paddatasetbenchmarkposeagnostic} & 4:33:43 & 66  \\
    SplatPose \cite{kruse2024SplatPosedetectposeagnostic} & 0:04:54 & 5 \\
    SplatPose+ & 0:03:04 & 0.034 \\
    \bottomrule
  \end{tabular}
\caption{Training and inference time on the MAD-SIM Dataset.}
\end{table}

\subsection{Robustness on Sparse-View Data} \label{section_sparse_view}
The MAD-Sim Dataset provides $210$ evenly distributed reference views for each object. In practice, the number of reference views might be limited. Therefore, the anomaly detection performance on sparse-view data is crucial in real-world applications. As a result, we also evaluated our method on randomly sampled partial training data ($20\%$, $40\%$, $60\%$, and $80\%$) to simulate sparse-view settings. Quantitatively, SplatPose+ achieves the highest AUROC and AUPRO scores given sparse-view training data under all settings. The difference is the most significant when using only $20\%$ of the training data because learning-based 3D representations (NeRF and 3DGS) will be blurry given limited training views, and, as a result, localizing with a learning-based 3D representation will produce inaccurate poses. On the other hand, a SfM model can still localized query poses well given $20\% \cdot 210 = 42$ reference views. Qualitatively, as illustrated in \cref{tab:spareview_viz}, SplatPose+ predicts the cleanest anomaly map for all three kinds of defects in the MAD-Sim dataset under all sparse-view settings. 

\begin{table}[!ht]
  \centering
  \begin{tabular}{ m{2cm} m{2cm} @{} m{2cm} @{} m{2cm} @{} | @{}m{2cm}@{} m{1.5cm} }
  Methods & \multicolumn{3}{ c }{Percentage of Reference Images Used} & \multicolumn{2}{ c }{Input \& } \\
     & 20\% & 60\% & 100\% & \multicolumn{2}{ c }{Ground Truth}  \\
    Ours & \includegraphics[width=2cm,valign=b]{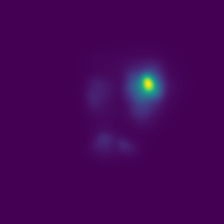} &
    \includegraphics[width=2cm,valign=b]{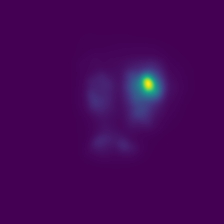} &
    \includegraphics[width=2cm,valign=b]{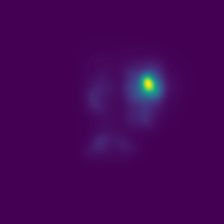} & 
    \includegraphics[width=2cm,valign=b]{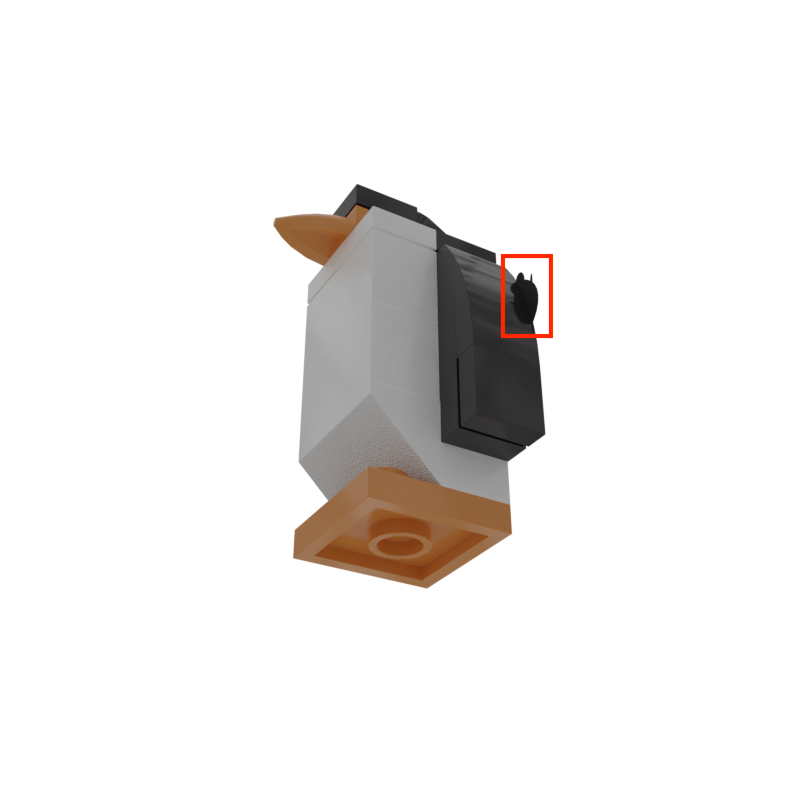} & Query (Burrs)\\
    SplatPose & \includegraphics[width=2cm,valign=b]{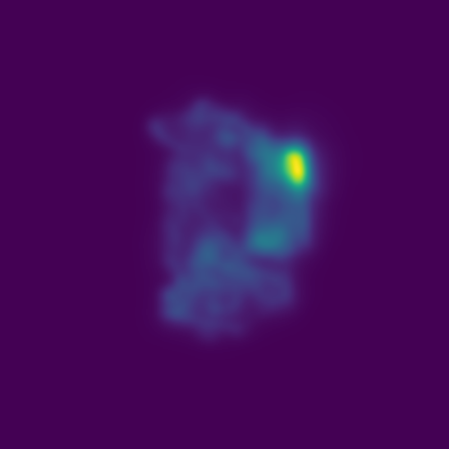} &
    \includegraphics[width=2cm,valign=b]{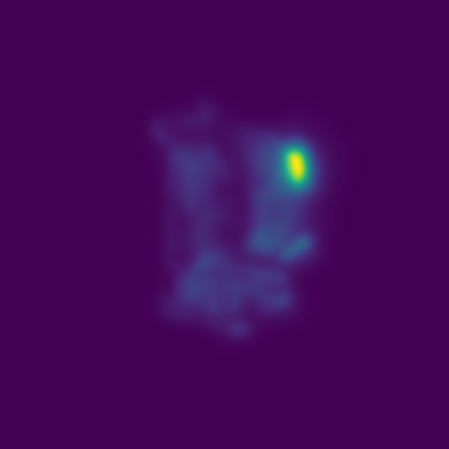} &
    \includegraphics[width=2cm,valign=b]{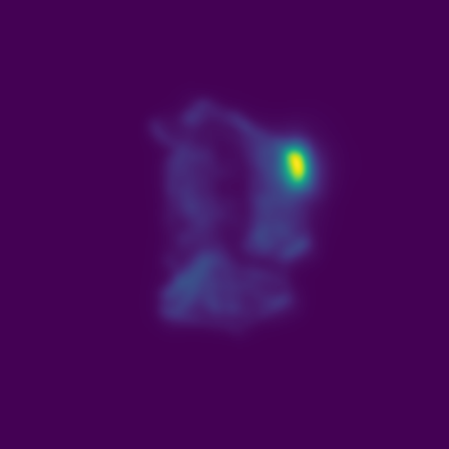} & 
    \includegraphics[width=2cm,valign=b]{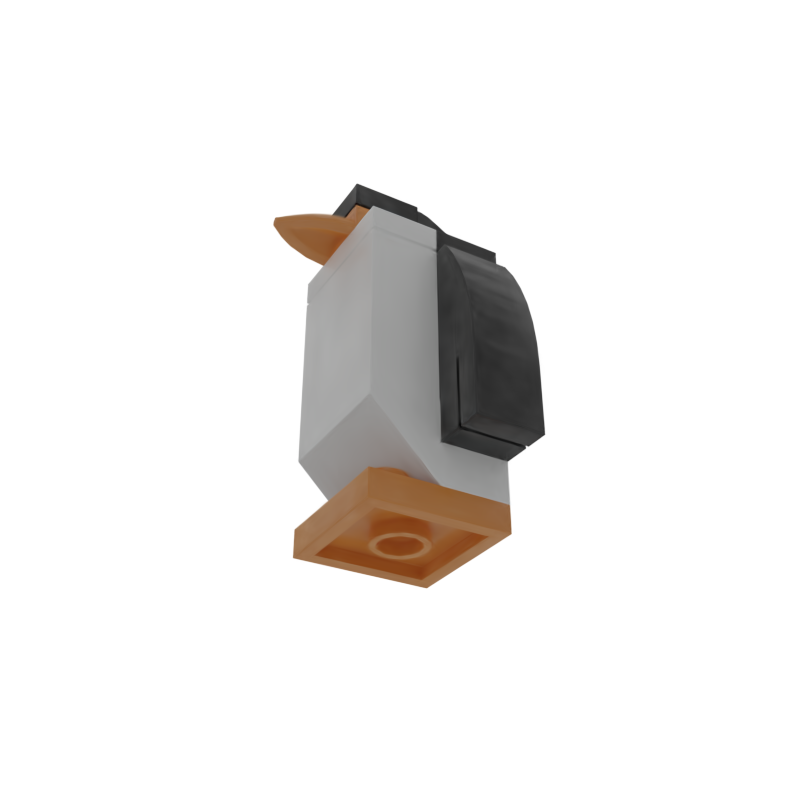} & Ours Pseudo Ref \\
    OmniposeAD & \includegraphics[width=2cm,valign=b]{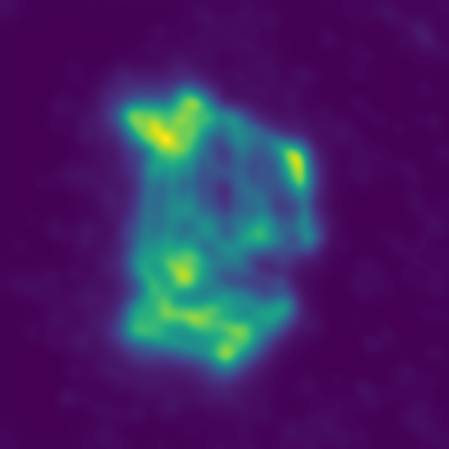} &
    \includegraphics[width=2cm,valign=b]{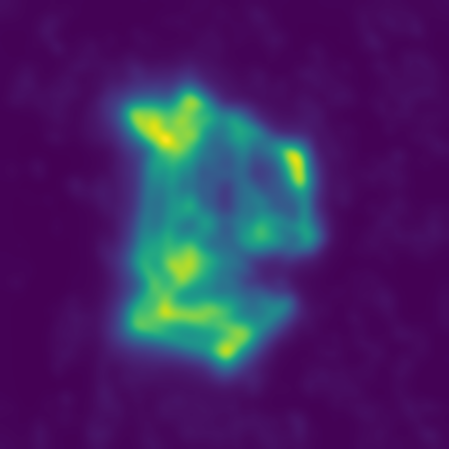} &
    \includegraphics[width=2cm,valign=b]{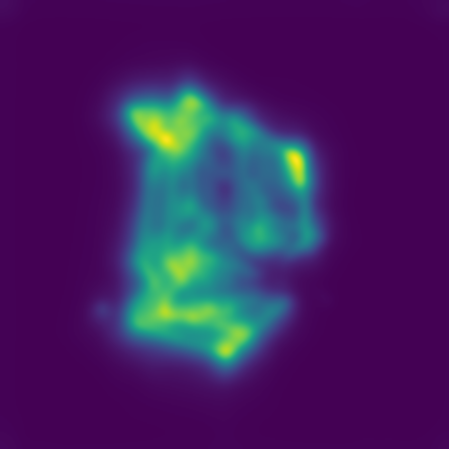} & 
    \includegraphics[width=2cm,valign=b]{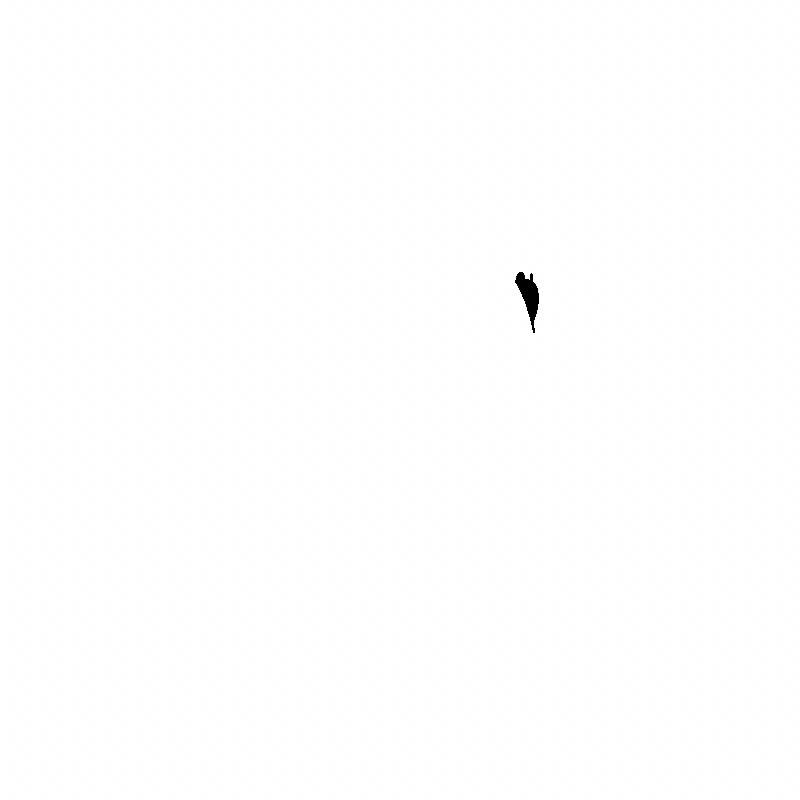} & Anomaly Mask \\
    \hline
    Ours & \includegraphics[width=2cm,valign=b]{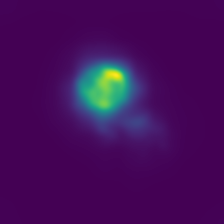} &
    \includegraphics[width=2cm,valign=b]{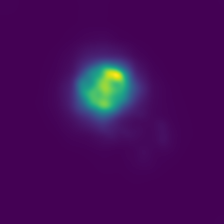} &
    \includegraphics[width=2cm,valign=b]{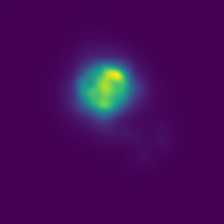} & 
    \includegraphics[width=2cm,valign=b]{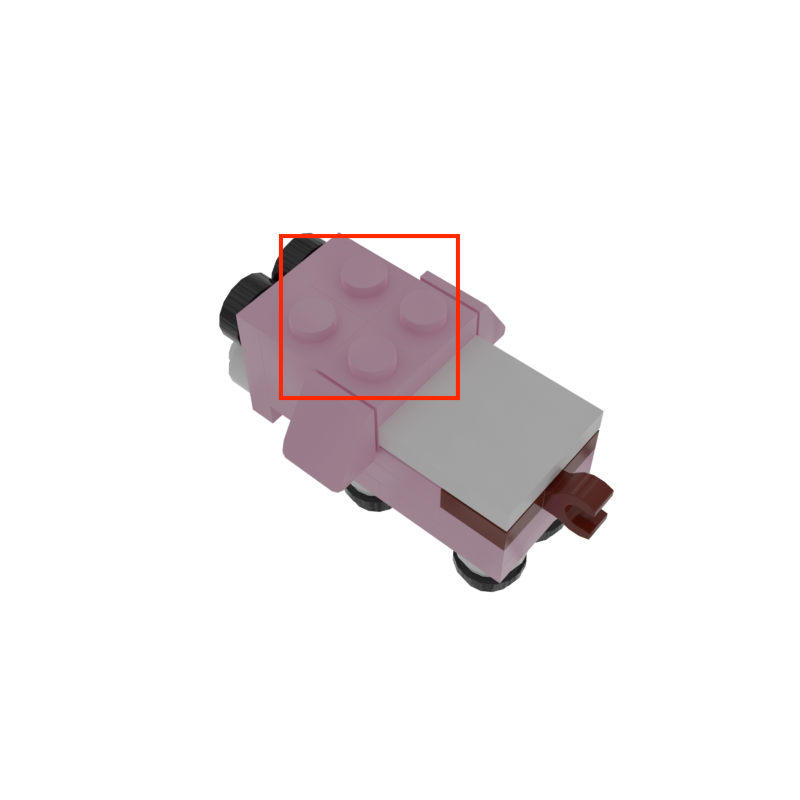} & Query (Missing) \\
    SplatPose & \includegraphics[width=2cm,valign=b]{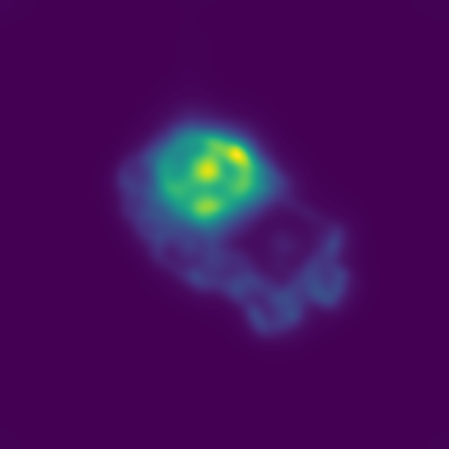} &
    \includegraphics[width=2cm,valign=b]{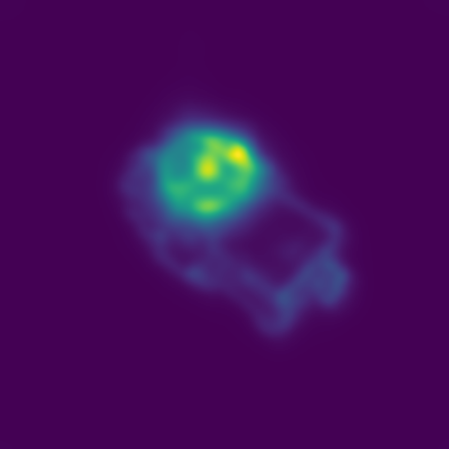} &
    \includegraphics[width=2cm,valign=b]{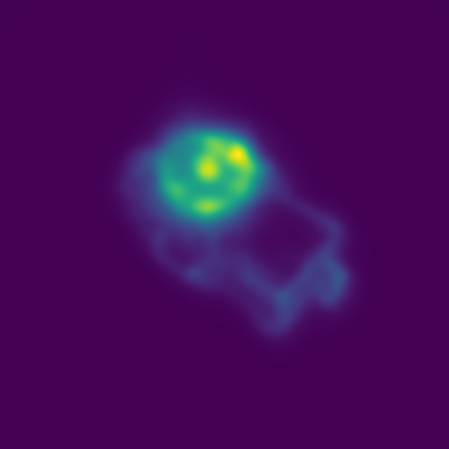} & 
    \includegraphics[width=2cm,valign=b]{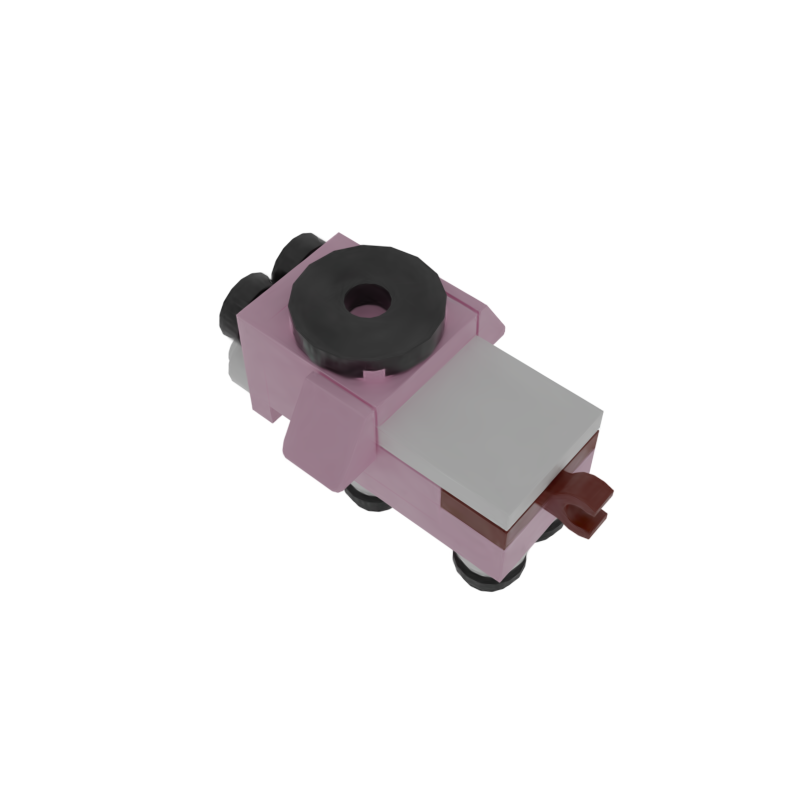} & Ours Pseudo Ref \\
    OmniposeAD & \includegraphics[width=2cm,valign=b]{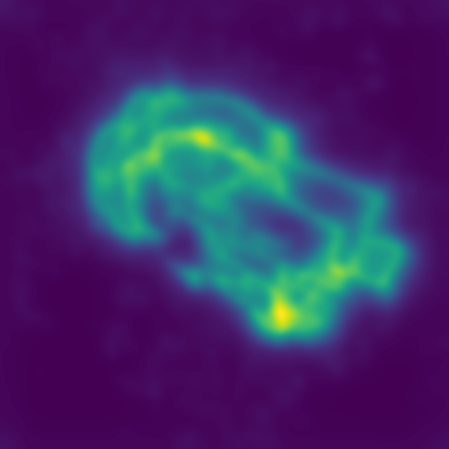} &
    \includegraphics[width=2cm,valign=b]{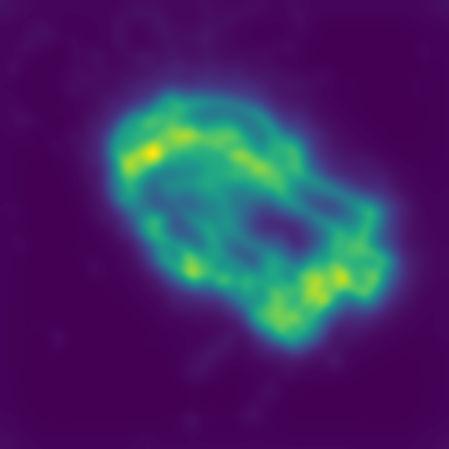} &
    \includegraphics[width=2cm,valign=b]{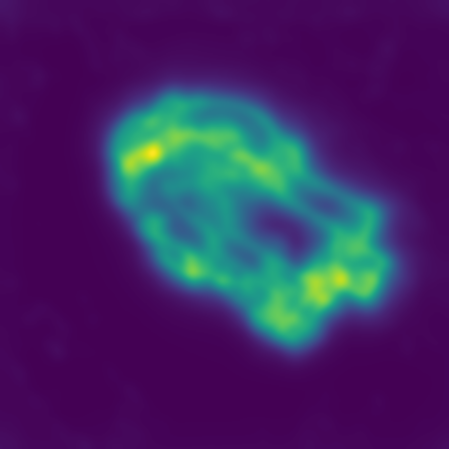} & 
    \includegraphics[width=2cm,valign=b]{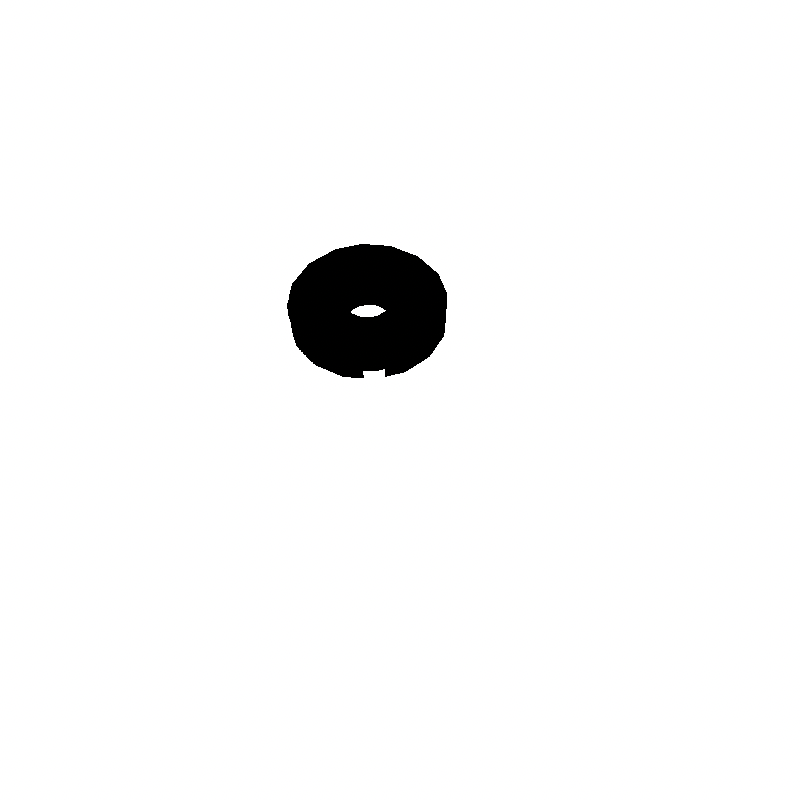} & Anomaly Mask \\
    \hline
    Ours & \includegraphics[width=2cm,valign=b]{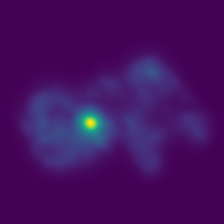} &
    \includegraphics[width=2cm,valign=b]{imgs/04Turtle_0.2.png} &
    \includegraphics[width=2cm,valign=b]{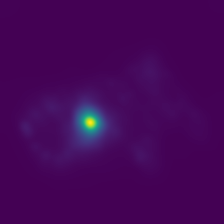} & 
    \includegraphics[width=2cm,valign=b]{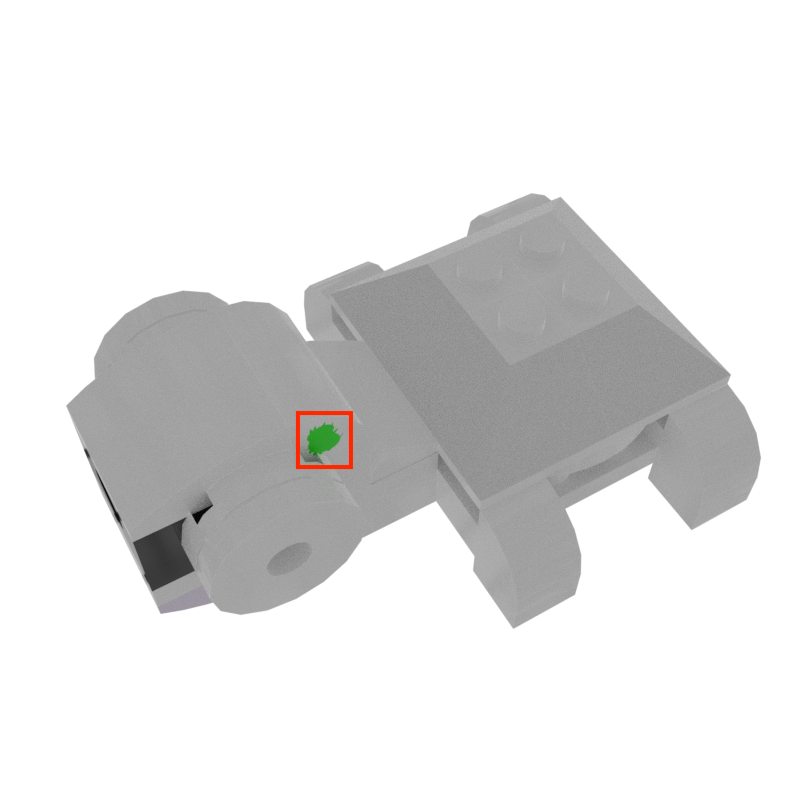} & Query (Stains) \\
    SplatPose & \includegraphics[width=2cm,valign=b]{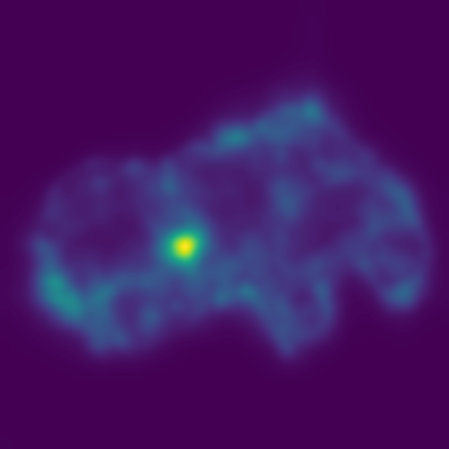} &
    \includegraphics[width=2cm,valign=b]{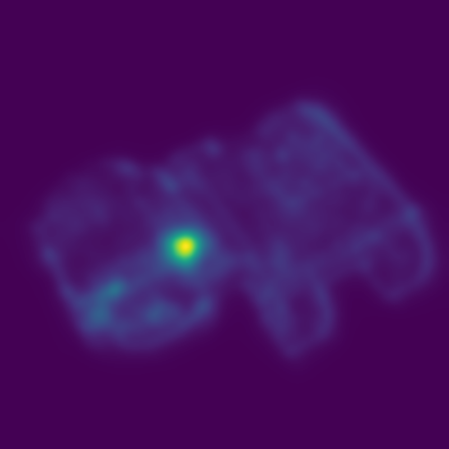} &
    \includegraphics[width=2cm,valign=b]{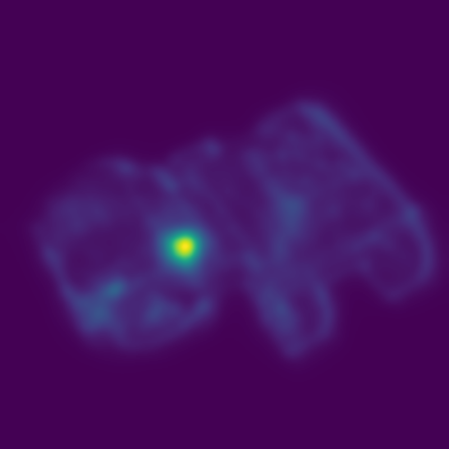} & 
    \includegraphics[width=2cm,valign=b]{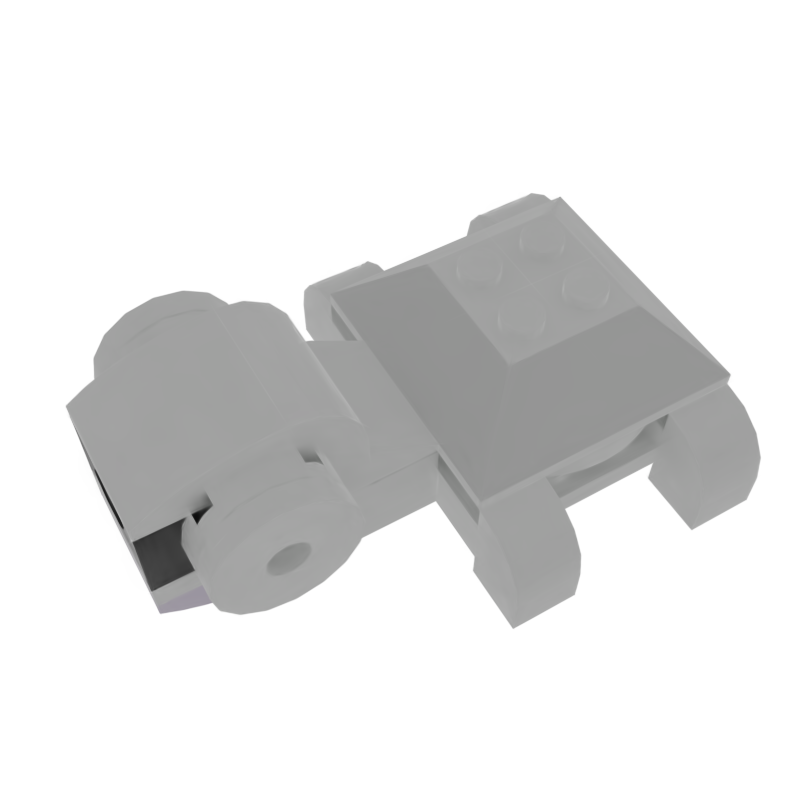} & Ours Pseudo Ref \\
    OmniposeAD & \includegraphics[width=2cm,valign=b]{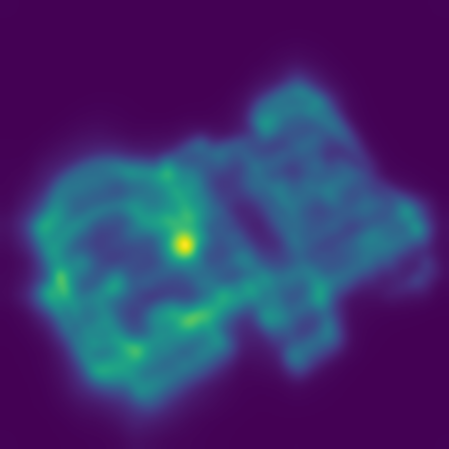} &
    \includegraphics[width=2cm,valign=b]{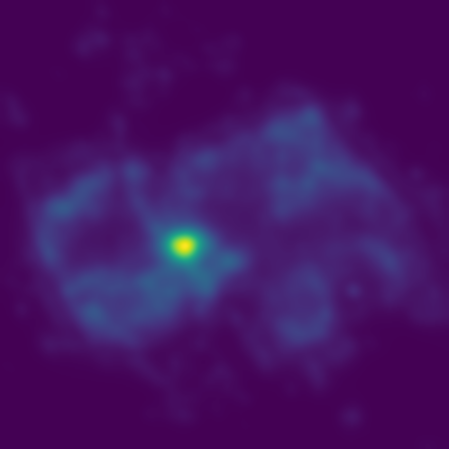} &
    \includegraphics[width=2cm,valign=b]{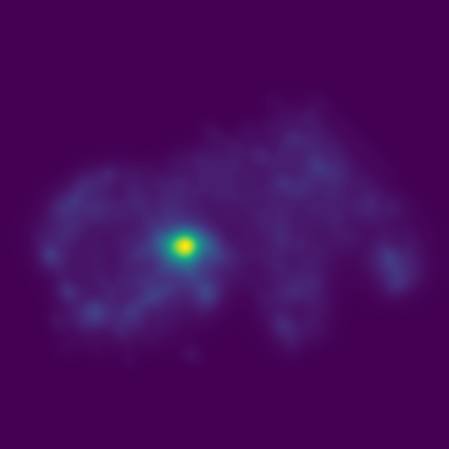} & 
    \includegraphics[width=2cm,valign=b]{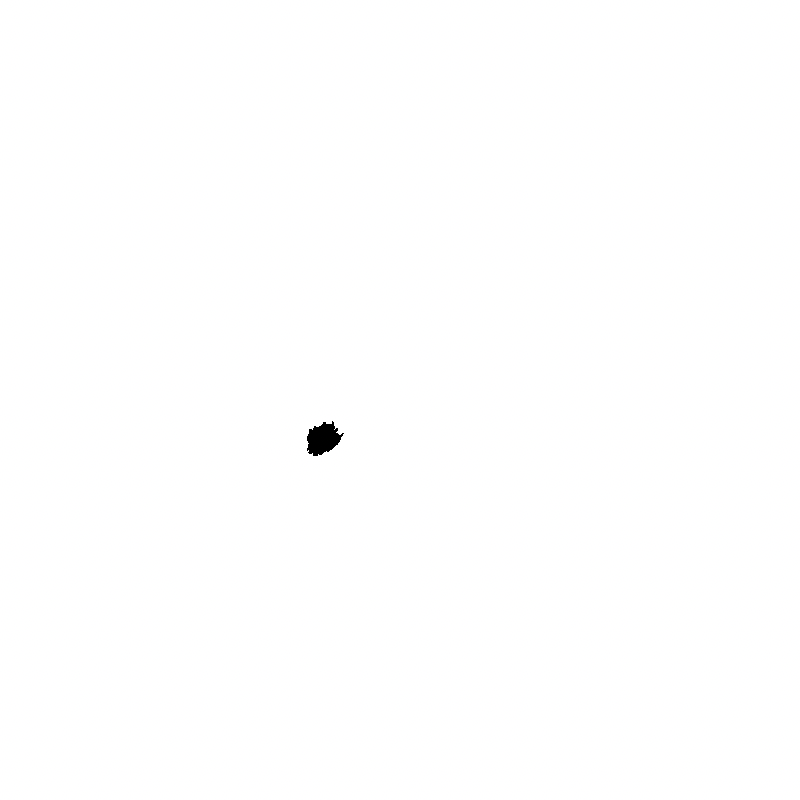} & Anomaly Mask \\
  \end{tabular}
\caption{Qualitative comparison of predicted anomaly maps under $3$ sparse-view settings. Ours Psuedo Ref is generated by SplatPose+ to help readers find the anomalies.}
\label{tab:spareview_viz}
\end{table}

\section{Conclusion}
In this paper, we present the first method that can infer in real-time for the Pose-agnostic 3D Anomaly Detection Task. We employ a hybrid representation comprised of a SfM model and a 3DGS model to encode a reference object. Given an un-posed query image, our method is able to localize it and find anomalies more accurately and $147$ times faster than the previous SOTA. Moreover, our method outperforms current methods qualitatively and quantitatively under the sparse-view settings. Future works can possibly incorporate more efficient 3DGS implementations, for example, Compact3D \cite{navaneet2024compact3dsmallerfastergaussian}, to further reduce the training and inference time. 
\clearpage

%
%
\bibliographystyle{splncs04}
\bibliography{main}
\end{document}